\documentclass[10pt, a4paper]{article}

\usepackage[final]{lrec-coling2024} 
\usepackage{amsmath}
\usepackage{amssymb}
\usepackage{graphicx}
\usepackage{times}
\usepackage{latexsym}
\usepackage{multirow}
\usepackage{tabularx}
\usepackage{algorithm}
\usepackage{algpseudocode}
\usepackage{subcaption}
\usepackage[T1]{fontenc}
\usepackage[utf8]{inputenc}
\usepackage{microtype}
\usepackage{inconsolata}
\usepackage{float}
\usepackage{hhline}
\usepackage{svg}

\title{DiffusionDialog: A Diffusion Model for Diverse Dialog Generation with Latent Space}

\name{
\begin{tabular}{@{}c@{}}
$\text{Jianxiang Xiang}^{1}$, 
$\text{Zhenhua Liu}^{1}$, 
$\text{Haodong Liu}^{2}$, \\
$\text{Yin Bai}^{2}$, 
$\text{Jia Cheng}^{2}$,  
$\text{Wenliang Chen}^{1}\sthanks{*Corresponding author}$ 
\end{tabular}
} 

\address{$^1$School of Computer Science and Technology, Soochow University, China \\
         $^2$Meituan \\
         \{jxxiang0720, zhliu0106\}@stu.suda.edu.cn, wlchen@suda.edu.cn \\
         \{liuhaodong05, baiyin, jia.cheng.sh\}@meituan.com \\}

\abstract{
In real-life conversations, the content is diverse, and there exists the one-to-many problem that requires diverse generation. Previous studies attempted to introduce discrete or Gaussian-based continuous latent variables to address the one-to-many problem, but the diversity is limited. Recently, diffusion models have made breakthroughs in computer vision, and some attempts have been made in natural language processing. In this paper, we propose DiffusionDialog, a novel approach to enhance the diversity of dialogue generation with the help of diffusion model. In our approach, we introduce continuous latent variables into the diffusion model. The problem of using latent variables in the dialog task is how to build both an effective prior of the latent space and an inferring process to obtain the proper latent given the context. By combining the encoder and latent-based diffusion model, we encode the response's latent representation in a continuous space as the prior, instead of fixed Gaussian distribution or simply discrete ones. We then infer the latent by denoising step by step with the diffusion model. The experimental results show that our model greatly enhances the diversity of dialog responses while maintaining coherence. Furthermore, in further analysis, we find that our diffusion model achieves high inference efficiency, which is the main challenge of applying diffusion models in natural language processing.
 \\ \newline  }

\begin{document}

\maketitleabstract

\section{Introduction}





\begin{figure}[t]
    \centering
    \includegraphics[width=8cm]{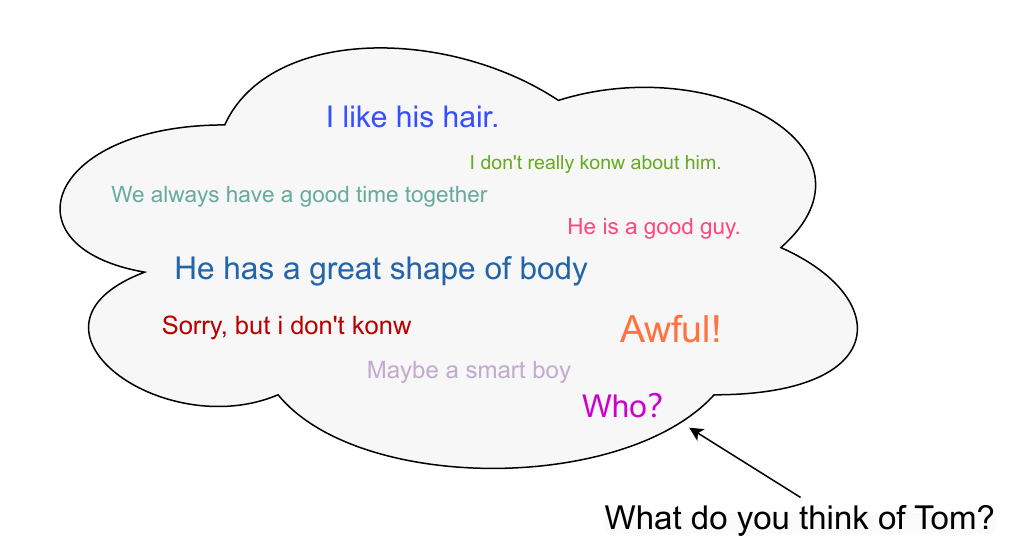}
    \caption{one to many problem in dialog generation.}
    \label{fig:One to many}
\end{figure}

Open-domain dialogue generation is a crucial component in dialogue systems. With the development of pre-trained language models, current models are capable of generating fluent and relevant dialogues\citep{radford2019language, raffel2020exploring}. However, there is still a lack of exploration in generating diverse responses, because there may be multiple appropriate responses when presented with a single context, and that's known as the one-to-many mapping problem, shown as figure \ref{fig:One to many}. To model the one-to-many relationship between dialog history and response, \citet{bao2019plato} introduce discrete latent variables, but the diversity of response is constrained by the categories of discrete latent variables, making it challenging to achieve fine-grained diversity generation. \citet{sun2021generating} and \citet{chen2022dialogved} introduce continuous latent variable which can relief the problem of the discrete latent variables, but the prior of the model is limited by the inflexible prior distribution, which cannot model the distribution of the response well.

As an alternative solution of one-to-many problem, we propose the integration of a diffusion model \citep{ho2020denoising}, which have shown its' superiority of generating high-quality and diverse results in the fields of image and audio generation \citep{dhariwal2021diffusion,ramesh2022hierarchical,rombach2022high,kong2020diffwave}. As for text-generation, DiffuSeq \citep{gong2022diffuseq} uses the Diffusion-LM \citep{li2022diffusion} structure for sequence-to-sequence tasks in a non-autoregressive manner, and both models perform diffusion operations in the embedding space. However, there are several important drawbacks. Firstly, the inference speed of the model will be greatly limited by the context length, especially in multi-turn dialogue scenarios where time consumption can be disastrous. Secondly, these models need to be trained from scratch and cannot take advantage of pre-trained language models. Some work has also attempted to combine diffusion models with latent variable. For example, LATENTOPS \cite{liu2022composable} applies diffusion models in latent space for controllable text generation tasks, this approach involves training multiple classifiers for different control requirements, and using the corresponding classifier to guide the inference of diffusion model in order to achieve controlled generation of text. However, as a complex conditional generation task, it is difficult to train classifiers to guide the latent inference process for dialogue generation.

In this work, we propose a structure that combines a latent-based diffusion model with a pre-trained language model to address the one-to-many modeling problem in multi-turn dialogues, called \textbf{DiffusionDialog}. DiffusionDialog integrates a encoder-decoder structured pre-trained language model Bart \citep{lewis2019bart} and a latent-based \citep{vaswani2017attention} diffusion model with transformer decoder structure. It performs inference of the diffusion model in the fixed-dimensional latent space, and combines the diffusion model with the language model for specific response generation. Instead of learning to approximate the fixed prior (e.g. Gaussian distribution) of the latent variable, our diffusion model learns a more flexible prior distribution from the encoder, enabling the generation of responses with finer-grained diversity. And due to the low-dimensional nature of the latent space, our diffusion model overcomes the slow inference speed issue which is the major problem of diffusion models.

The contributions of this paper can be summarized as follows:
\begin{enumerate}
    \item We propose a novel approach to address the one-to-many problem in dialogue using a combination of a latent-based diffusion model and a pre-trained language model.
    \item To the best of our knowledge, our work is the first to apply a latent diffusion model to dialog generation. By reasoning in the latent space, the inference efficiency of our diffusion model is significantly improved.
    \item Through comparative experiments, we demonstrate the effectiveness of our model, which can generate responses that are rich in diversity while ensuring fluency and coherence.
\end{enumerate}

\section{Background}
\subsection{Dialog Generation with Latent Variable}
The objective of dialog system is to estimate the conditional distribution $p(x|c)$. Let $d=[u_{1}, ..., u_{k}]$ denote a dialogue comprising of k utterances. Each utterance is represented by $u_{i}=[w_{1}, ..., w_{|ui|}]$, where $w_{n}$ refers to the $n$-th word in $u_{i}$. Additionally, we define $c=[u_{1}, ..., u_{k-1}]$ as the dialogue context, which constitutes the $k-1$ historical utterances, and $x=u_{k}$ as the response, which denotes the next utterance in the dialogue.

Finding a direct connection between the discrete token sequences $x$ and $c$ can be challenging. To address this issue, we propose the use of a continuous latent variable $z$, which serves as a high-level representation of the response. In this two-step response generation process, we first sample a latent variable $z$ from a distribution $p_{\theta}(z|c)$ that resides in a latent space $\mathcal{Z}$. Subsequently, we decode the response $x$ from $z$ and $c$ as $p_{\theta}(x|z, c)$.And this process can be estimated as
\begin{equation}
\label{E1}
p_{\theta}(x|c) = \int_{z}p_{\theta}(z|c)p_{\theta}(x|z, c)d_{z}.
\end{equation}

Since the optimal $z$ is intractable, we optimize the posterior distribution of $z$ as $q_{\phi}(z|x)$ considering the $x$. And we approximate the posterior with the prior distribution $p_{\theta}(z|c)$,
\begin{equation}
\begin{array}{l}
\label{E2}
\log p_{\theta}(x|c) = \log \int_{z}q_{\phi}(z|x)p_{\theta}(x|z, c) \\ \geq E_{z \sim q_{\phi}(z|x)}[\log p_{\theta}(x|z, c)] \\- KL(q_{\phi}(z|x), p_{\theta}(z|c)).
\end{array}
\end{equation}

\subsection{Diffusion Model in Latent Space}
Diffusion model is designed to operate in fixed and continuous domain, consisting forward and reverse processes. In this work, we perform forward and reverse process in learned latent space representing the high-level semantic of response. Suppose posterior as $z_{0} \sim q_{\phi}(z|x)$, in the forward process, $z_{0}$ is corrupted with standard Gaussian noise in large amount of step, forming a Markov chain of $z_{0}, z_{1},...,z_{T}$, with ${z}_{T} \sim \mathcal{N}(0, I)$:
\begin{equation}
\label{E3}
q(z_{t}|z_{t-1}) = \mathcal{N}(z_{t};\sqrt{1-\beta_{t}}z_{t-1},\beta_{t}I),
\end{equation}
where $\beta_{t} \in (0,1)$ controls the scale of the noise in a single step.

In the reverse progress, diffusion model learn to reconstruct $z_{0}$ from $z_{T}$ by learning $p_{\theta}(z_{t-1}|z_{t}) = \mathcal{N}(z_{t-1};\mu_{\theta}(z_{t},t),\Sigma_{\theta}(z_{t},t))$, Since the $q(z_{t-1}|z_{t},z_{0})$ has a closed form,the canonical objective is the variational lower bound of $\log p_{\theta}(z_{0})$,
\begin{equation}
        \label{E4}
        \begin{array}{l}
\mathcal{L}_{\mathrm{vlb}}=\mathbb{E}_{q}\left[D_{\mathrm{KL}}\left(q\left(z_{T} \mid z_{0}\right) \| p_{\theta}\left(z_{T}\right)\right)\right] \\
+\mathbb{E}_{q}\left[\sum_{t=2}^{T} D_{\mathrm{KL}}\left(q\left(z_{t-1} \mid z_{t}, z_{0}\right) \| p_{\theta}\left(z_{t-1} \mid z_{t}, t\right)\right)\right] \\
-\log p_{\theta}\left(z_{0}\mid z_{1}\right).
        \end{array}
\end{equation}

To promote stability in training, we take advantage of the simplified objective proposed by \citeauthor{ho2020denoising} as $\mathcal{L}_{\text {simple}}$,
\begin{equation}
    \label{E5}
    \mathcal{L}_{\text {simple}}(z_{0})=\sum_{t=1}^{T} \underset{q(z_{t}| z_{0})}{\mathbb{E}}\|\mu_{\theta}(z_{t}, t)-\hat{\mu}\left(z_{t}, z_{0}\right)\|^{2},
\end{equation}
where $\hat{\mu}(z_{t}, z_{0})$ refers to $q(z_{t-1}|z_{t},z_{0})$, and $\mu_{\theta}(z_{t}, z_{0})$ is learned by diffusion model.

\section{DiffusionDialog}
\subsection{Model Architecture}

Our model introduces a hierarchical generation process with latent variable. Firstly it obtains latent variable reflecting the semantic of response from the context and then generate the response considering the latent variable and the context (Equation \ref{E1}), thus the response generation involves three key components: the dialogue context $c$, the response $r$, and the latent variable $z$.

\begin{figure*}[t]
    \centering
    \includegraphics[width=16cm]{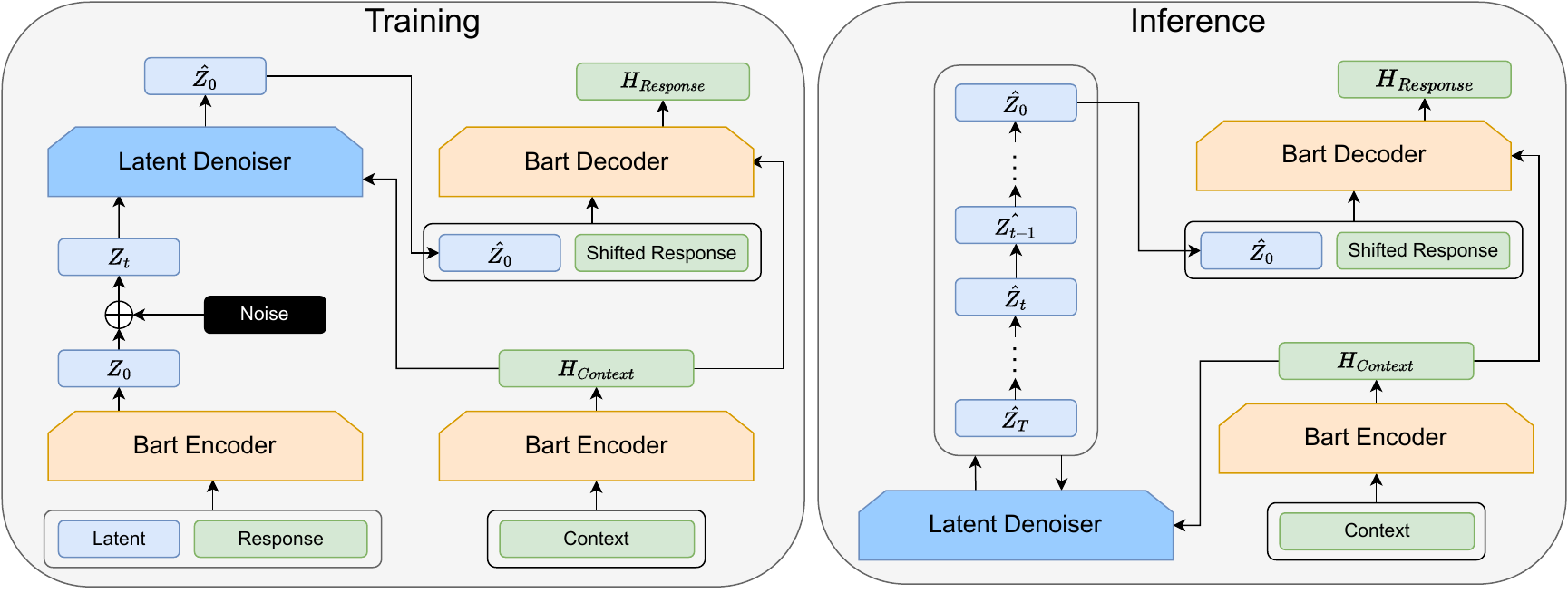}
    \caption{frame work of DiffusionDialog.}
    \label{fig:DiffusionDialog}
\end{figure*}

We combines encoder-decoder structured pre-trained language model Bart with a latent-based diffusion model to handle the two-stage generation, the figure \ref{fig:DiffusionDialog} illustrates our model, and we explain our model by illustrating the function of each part of the model.

\subsubsection{Bart Encoder}
The bart encoder plays a dual role in our model, encoding both the contex and the latent variables.

For context, following the PLATO, in addition to token and position embeddings, it also incorporates turn embeddings to align with the context turn number, and role embeddings to align with the speaker's role. As a result, the final embedding input of the context is the sum of corresponding token, turn, role, and position embeddings.

For latent variables, since the priors are untraceable, bart encoder learns the priors of the latent variable $q_{\phi}(z|x)$ which represents the high-level semantic information about the response.


To connect the latent space, we concatenate a special token in front of the response to encode the semantic information of the response. We refer to this special token as latent toke. Therefore, the input format for latent variable encoding is $[l, w_{1}^{x}, w_{2}^{x} ..., w_{n}^{x}]$, $n$ refers to the length of response $x$.

We append a multilayer perceptron to obtain a representation of the posterior distribution $z_0 \sim q_{\phi}(z|x)$ :
\begin{equation}
    \label{E6}
    z_0=MLP(h_{[L]}),
\end{equation}
where $h_{[L]}\in\mathbb{R}^{d}$ refers to the final hidden state of the latent token.


\subsubsection{Latent Diffusion Denoiser}
After obtaining $z_0$ from the bart encoder, we sample a time step $t \in [1,T]$ uniformly and add noise to the latent variable according to Equation \ref{E3}, resulting in a noised latent $z_{t}$. The latent diffusion denoiser is trained to denoise the latent. It adopts the structure of a transformer decoder, taking the noised latent variable as inputs and incorporates the context hidden state with cross-attention mechanism, and a timestep embedding is also added before the first Transformer block to inform the model of the current timestep,
\begin{equation}
    \Tilde{z_0} = Denoiser(z_{t}, e_t, h_c),
\end{equation}
where $e_t$ refers to the embedding of the timestep $t$. Since the context hidden state is fixed during inference, the inference time required for the diffusion model is short.

\subsubsection{Bart Decoder}

To guide the response generation of the decoder using latent variables, we adopt the memory scheme from \textbf{OPTIMUS} \citep{li2020optimus}. Specifically, we project the latent variable $z$ as a key-value pair and concatenate them to the left of the token hidden state to introduce the latent variable into the decoder.
\begin{small}
\begin{align*}
H^{(l+1)} = MultiHead(H^{(l)},h_{Mem}^{(l)} \oplus H^{(l)}, h_{Mem}^{(l)} \oplus H^{(l)}),
\end{align*}
\end{small}
where $H^{(l)}$ refers to the token hidden state of the $l$-th layer, and $h_{Mem}^{(l)}$ is calculated as:
\begin{equation}
\label{E7}
h_{Mem}^{(l)}=\left[
\begin{matrix}
z_{key} \\
z_{value}
\end{matrix}
\right] = W_{M}^{l}\,z,
\end{equation}
where $W_{M}^{l}\in \mathbb{R}^{d \times 2d}$ is a weight matrix.

\subsection{Training}
During our training process for dialogue generation, we utilize three different loss functions: negative log-likelihood (NLL) loss, bag-of-words (BOW) loss, and latent denoising (LD) loss. Detailed descriptions will be provided in this section.

\subsubsection{Response semantic capture}
To enable the latent variable to capture the overall semantic information of the response, we adopt the bag-of-words (BOW)\citep{zhao2017learning} loss, which is used to enable the latent variable to predict the tokens in the response in a non-autoregressive manner.
\begin{equation}
\label{E8}
	\begin{aligned}
	\mathcal{L}_{BOW}&=-\mathbb{E}_{z_0\sim q_{\phi}(z|r)}  \sum_{n=1}^N ~ \log p(r_t|z_0)\\
	&=-\mathbb{E}_{z_0\sim q_{\phi}(z|r)}  \sum_{n=1}^N ~ \log \frac{e^{f_{r_n}}}{\sum_{v\in V} e^{f_v}}~.
	\end{aligned}
\end{equation}

The symbol $V$ refers to the entire vocabulary. The function $f$ attempts to non-autoregressively predict the words that make up the target response.
\begin{equation}
f=\operatorname{softmax}\left(W_{2} h_{z}+b_{2}\right) \in \mathbb{R}^{|V|}.
\end{equation}

In the given equation, $h_{z}$ represents the hidden state of the latent variable, while $|V|$ denotes the size of the vocabulary. The estimated probability of word $r_n$ is denoted by $f_{r_n}$. BOW loss disregards the word order and compels the latent variable to capture the overall information of the target response.

\subsubsection{Latent Denoising}
For each training step, we sample a time step $t$ and obtain $z_t$ referring to Equation \ref{E3}. 
To better capture the semantic information of the latent variables, our diffusion model predicts $z_0$ directly instead of $z_{t-1}$ given $z_t$, denoted as $\mathcal{L}_{z_{0} \text {-simple }}$, a variant of $\mathcal{L}_{\text {simple }}$ in Equation \ref{E5}: 
\begin{equation}
\begin{aligned}
\mathcal{L}_{z_{0} \text {-simple }}\left(z_{0}\right)=\sum_{t=1}^{T} \mathbb{E}_{z_{t}}\left\|p\left(z_{t}, c, t\right)-z_{0}\right\|^{2}.
\end{aligned}
\end{equation}
where our latent diffusion denoiser $p\left(z_{t}, h_{c}, t\right)$ predicts $z_{0}$ directly.

Thus at each time step, the loss of latent denoising is:
\begin{equation}
\begin{aligned}
\mathcal{L}_{LD}&=\|p\left(z_{t}, t, c\right)-z_{0}\|^{2}.
\end{aligned}
\end{equation}

\subsubsection{Response Generation}
In our model, the response is generated by conditioning on both the latent variable and the context. To train the response generation we adopt the commonly used NLL loss,
\begin{equation}
\begin{aligned}
\mathcal{L}_{N L L} & =-\mathbb{E}_{\Tilde{z_0} \sim p(z \mid c, z_t, t)} \log p(r \mid c, \Tilde{z_0}) \\
& =-\mathbb{E}_{\Tilde{z_0} \sim p(z \mid c, z_t, t)} \sum_{n=1}^{N} \log p\left(r_{t} \mid c, \Tilde{z_0}, r_{<t}\right).
\end{aligned}
\end{equation}

Note that $\Tilde{z_0}$ is the posterior distribution predicted by the latent diffusion denoiser, we adopt this approach to reduce the gap between training and inference. In order to optimize the NLL loss, the denoiser's prediction needs to not only be close to the prior distribution $z_{0}$ in the spatial domain, but also approximate the response in the semantic domain.

In summary, our model aims to minimize the overall objective function, which is defined as the integrated loss:
\begin{equation}
\mathcal{L} = \mathcal{L}_{NLL} + \mathcal{L}_{BOW} + \mathcal{L}_{LD}.
\end{equation}

\subsection{Inference}

The inference in our model consists of two stages. Firstly, starting from a Gaussian noise, the latent diffusion denoiser performs multiple rounds of inference to denoise the latent variable and obtain the final semantic representation $z_0$, conditioned on the hidden state of the context which is encoded by the encoder. Then the response generator generates the final response in an auto-regressive manner, conditioned on both $z_0$ and the context hidden state.

For ease of displaying the training and inference process of our model, we outline our approach in Figure \ref{fig:algorithms}.

\begin{figure}[htb]
    \centering
    \begin{minipage}{0.45\textwidth}
        \centering
        \begin{algorithm}[H]
            \caption{Training}
            \begin{algorithmic}[1]
                \Statex \textbf{Input:} a dialog corpus $\mathcal{D}$=$\{(c_i, r_i)\}_{i=1}^{|\mathcal{D}|}$
                \State \textbf{repeat}
                \State sample context and response (c, r) from $\mathcal{D}$
                \State $h_{c} = \mathrm{Encoder}(c)$
                \State $z_{0} \sim q_{\phi}(z|r) = \mathrm{Encoder}([l; x])[0]$
                \State $\mathcal{L}_{BOW}=-\sum_{n=1}^N ~ \log p(r_t|z_{0})$
                \State $t \sim \mathrm{Uniform}(\{1, \dotsc, T\})$
                \State $\epsilon \sim \mathcal{N}(0, I)$
                \State $z_{t} = \sqrt{\Bar{\alpha}_t} z_{0} + \sqrt{1-\Bar{\alpha}_t} \epsilon$
                \State $\Tilde{z_{0}} = \mathrm{Denoiser}(z_{t}, h_{c}, t)$
                \State $\mathcal{L}_{LD}=-\|\Tilde{z_{0}}-z_{0}\|^{2}$
                \State $\mathcal{L}_{N L L} =-\sum_{n=1}^{N} \log p\left(r_{t} \mid c, \Tilde{z_{0}}, r_{<t}\right)$
                \State Take gradient descent step on
                \Statex $\qquad \nabla_\theta [\mathcal{L} = \mathcal{L}_{LD} + \mathcal{L}_{N L L} + \mathcal{L}_{BOW}]$
                \State \textbf{until} converged
            \end{algorithmic}
        \end{algorithm}
    \end{minipage}
    \hfill
    \begin{minipage}{0.45\textwidth}
        \centering
        \begin{algorithm}[H]
            \caption{Inference}
            \begin{algorithmic}[1]
                \State $h_{c} = \mathrm{Encoder}(c)$
                \State $\Tilde{z_{T}} \sim \mathcal{N}(0, I)$
                \State \textbf{for} $t=T, \dotsc, 1$ \textbf{do}
                \State $\quad \Tilde{z_{0}} = \mathrm{Denoiser}(\Tilde{z_{t}}, h_{c}, t)$
                \State $\quad \epsilon \sim \mathcal{N}(0, I)$
                \State $\quad \Tilde{z_{t-1}} = \sqrt{\Bar{\alpha}_{t-1}} \Tilde{z_{0}} + \sqrt{1-\Bar{\alpha}_{t-1}} \epsilon$
                \State \textbf{end for}
                \State response $\Tilde{r} = Decoder(\Tilde{z_{0}}, h_{c})$
            \end{algorithmic}
        \end{algorithm}
    \end{minipage}
    \caption{The training and inference algorithm of DiffusionDialog.}
    \label{fig:algorithms}
\end{figure}



\section{Experiments}

\subsection{Experimental Setup}

\subsubsection{Datasets and Evaluation}
Following PLATO\citep{bao2019plato},we evaluate the performance of our model on two commonly used public dialog datasets.

\textbf{DailyDialog}\citep{li2017dailydialog} is a high-quality conversational dataset that primarily focuses on daily dialogues.

\textbf{Persona-Chat}\citep{zhang2018personalizing} is sourced from authentic conversations between human annotators who are randomly matched and assigned a given persona information. Paired annotators engage in natural conversation and attempt to know each other better throughout the dialogue.

Table \ref{tab:dataset} summarizes the descriptions and statistics of these datasets. In DailyDialog, 12.1\% of the development set and 13.0\% of the test set appeared in the training set, indicating the presence of data leakage, while no such issue is observed in PersonaChat.
\begin{table}[t]
\centering
\scalebox{1.0}{
\begin{tabular}{c|c|c}
\hline
      & DailyDialog                                                              & PersonaChat                                                               \\ \hline
train & \begin{tabular}[c]{@{}c@{}}76052 samples\\ \textbackslash{}\end{tabular} & \begin{tabular}[c]{@{}c@{}}122499 samples\\ \textbackslash{}\end{tabular} \\ \hline
dev   & \begin{tabular}[c]{@{}c@{}}7069 samples\\ 12.1\% overlap\end{tabular}     & \begin{tabular}[c]{@{}c@{}}14602 samples\\ \textbackslash{}\end{tabular}  \\ \hline
test  & \begin{tabular}[c]{@{}c@{}}6740 samples\\ 13.0\% overlap\end{tabular}     & \begin{tabular}[c]{@{}c@{}}14056 samples\\ \textbackslash{}\end{tabular}  \\ \hline
\end{tabular}}
\caption{Summary of datasets used in the experiments, overlap means percentage of data leaks.}
\label{tab:dataset}
\end{table}

\begin{table*}[ht]
\centering
\scalebox{1.0}{
\begin{tabular}{l|cccc}
\hline
\multicolumn{1}{c|}{\multirow{2}{*}{Model}} & \multicolumn{4}{c}{DailyDialog}           \\
\multicolumn{1}{c|}{}     & BLEU-1 & BLEU-2 & Distinct-1 & Distinct-2 \\ \hline
Seq2Seq~\cite{vinyals2015neural}                & 0.336  & 0.238  & 0.030      & 0.128      \\
iVAE\_MI~\cite{fang2019implicit}               & 0.309  & 0.249  & 0.029      & 0.250      \\
PLATO w/o latent$^{\dagger}$ \cite{bao2019plato}       & 0.405  & 0.322  & 0.046      & 0.246      \\
PLATO$^{\dagger}$ \cite{bao2019plato}                  & 0.397  & 0.311  & 0.054      & 0.291      \\
DialogVED$^{\ddagger}$ \cite{chen2022dialogved}              &  \underline{0.481}  &  \underline{0.421}  & 0.042      & 0.232      \\ \hline
Our w/o Latent         & \textbf{0.406}  & \textbf{0.371}  & 0.046      & 0.217      \\
Our Method             & 0.348  & 0.318  & \underline{\textbf{0.072}}      & \underline{\textbf{0.372}}      \\
Our Method Upper Bound & 0.471  & 0.424  & 0.063      & 0.348      \\ \hline
\hline
\multicolumn{1}{c|}{\multirow{2}{*}{Model}} & \multicolumn{4}{c}{PersonaChat}           \\
\multicolumn{1}{c|}{}  & BLEU-1 & BLEU-2 & Distinct-1 & Distinct-2 \\ \hline
Seq2Seq~\cite{vinyals2015neural}                & 0.448  & 0.353  & 0.004      & 0.016      \\
LIC~\cite{golovanov2019large}                    & 0.405  & 0.320  & 0.019      & 0.113      \\
PLATO w/o latent$^{\dagger}$ \cite{bao2019plato}       & \textbf{0.458}  & 0.357  & 0.012      & 0.064      \\
PLATO$^{\dagger}$ \cite{bao2019plato}                  & 0.406  & 0.315  & 0.021      & 0.121      \\
DialogVED$^{\ddagger}$ \cite{chen2022dialogved}              &  \underline{0.482}  &  \underline{0.399}  & 0.015      & 0.094      \\ \hline
Our w/o Latent         & 0.410  & \textbf{0.359}  & 0.013      & 0.056      \\
Our Method             & 0.383  & 0.329  & \underline{\textbf{0.031}}      & \underline{\textbf{0.177}}      \\
Our Method Upper Bound & 0.472  & 0.411  & 0.026      & 0.166      \\ \hline
\end{tabular}}
\caption{Experimental results on DailyDialog and PersonaChat with automatic evaluations. PLATO and DialogVED is pretrained with large dialog corpus, and DialogVED is based on Bart Large. The best values are \underline{underlined}, and the best results with base-PLMs is written in bold.}
\label{tab:dailidialog}
\end{table*}

\begin{table*}[ht]
\centering
\begin{tabular}{c|l}
\hline
\multirow{3}{*}{context}    & {[}P1{]}It's a lovely day out today, isn't it?                                       \\
                            & {[}P2{]}It's beautiful. Enjoy it while it lasts. It's supposed to get cold tomorrow. \\
                            & {[}P1{]}What's the weather forecast for tomorrow?                                    \\ \hline
our w/o Latent              & It's supposed to snow                                                                \\ \hline
\multirow{8}{*}{our Method} & Cloudy in the morning and overcast in the afternoon                                  \\ \cline{2-2} 
                            & The weatherman says it's going to snow tomorrow                                      \\ \cline{2-2} 
                            & It's supposed to snow in the morning and in the afternoon. Is that possible?         \\ \cline{2-2} 
                            & The weatherman says it's supposed to snow all day.                                   \\ \cline{2-2} 
                            & The weatherman says a storm is coming.                                               \\ \cline{2-2} 
                            & Yep. It's supposed to snow in the morning.                                           \\ \cline{2-2} 
                            & Cloudy with a chance of showers.                                                     \\ \cline{2-2} 
                            & Dreadful. It's supposed to snow tomorrow.                                            \\ \hline
\hline
\multirow{3}{*}{context}    & {[}P1{]} Good morning, sir. Is there a bank near here?           \\
                            & {[}P2{]} There is one. 5 blocks away from here?                  \\
                            & {[}P3{]} Well, that's too far. Can you change some money for me? \\ \hline
our w/o Latent              & Yes, Please.                                                     \\ \hline
\multirow{6}{*}{our Method} & Yes, Please wait for a moment.                                   \\ \cline{2-2} 
                            & Yes, madam. I am sure you can. The interest rate is very high.   \\ \cline{2-2} 
                            & What's your account number and your PIN number?                  \\ \cline{2-2} 
                            & Yes, Madam. Can I help you?                                      \\ \cline{2-2} 
                            & How can I help you?                                              \\ \cline{2-2} 
                            & Yes, certainly.My name is John Sandals.                          \\ \hline
\end{tabular}
\caption{Examples of response generation with our model.}
\label{tab:case}
\label{tab:example}
\end{table*}

For evaluation, we mainly evaluate our model on fluency and diversity. We adopt the same metrics used in PLATO, which is widely used:

\textbf{BLEU-1/2}\citep{papineni2002bleu} which measures the coherence of generated response to the given context by calculating the 1/2-grams overlapping between the generated response and references.

\textbf{Distinct-1/2}\citep{li2015diversity} which measures the diversity of generated response by calculating the number of unique 1/2-grams divided by the total number of generated words.


\subsubsection{Compared Baselines}

In our experiments, the following models were selected as our baselines.

\textbf{Seq2Seq}\citep{vinyals2015neural} is a sequence-to-sequence model with attention. \textbf{IVAE$_{\text {MI}}$}\citep{fang2019implicit} is also a sequence-to-sequence model with implicit deep latent variable that employs a Variational Autoencoder to improve the quality of latent representations and generate diverse responses. \textbf{LIC}\citep{golovanov2019large} is a transformer-based generative model fine-tuned on GPT, which has demonstrated remarkable performance in the ConvAI2 challenge. \textbf{PLATO}\citep{bao2019plato} employs a discrete latent variable to address the one-to-many problem, showing high performance in both response fluency and diversity. \textbf{DialogVED}\citep{chen2022dialogved} introduces continuous latent variables with VAE model into the enhanced encoder-decoder pre-training framework to increase the relevance and diversity of responses. Both of PLATO and DialogVED address the one-to-many problem in dialogue tasks and are the main objects of comparison in our study.

To accurately evaluate the impact of our latent variable with diffusion model, we compare our model to the version without the latent.

\subsubsection{Model Configuration}
Our model consists of two parts: one is a encoder-decoder structure Transformer model Bart-base, which is composed of 6 layers of encoder and 6 layers of decoder. The other part of our model is a latent denoiser, which is a structure of 6 layers of transformer decoder with latent token embedding and 128-dimensional time step embedding. Our diffusion steps $T=2,000$ and noise schedule is square-root schedule. Our maximum context squence length is 256 and our maximum response sequence length 128, The model uses the BPE tokenization\citep{sennrich2015neural} which is commonly used.

During training, We use Adamw optimizer\citep{loshchilov2017decoupled} with a learning rate of $1 \times 10^{-4} $, the batch size is 128, We also adopt a warmup strategy where we linearly increase the learning rate from initial learning rate $1 \times 10^{-7} $, the totall training steps for DailyDialog is 10000, and for PersonaChat is 20000. We select the checkpoint with the lowest validation loss for inference. The experiment is carried out on one single 1080Ti GPU.

\subsection{Main Results}

Table \ref{tab:dailidialog} summarizes the experimental results on Persona-Chat and Daily Dialog.

Note that both the PLATO and DialogVED models have been pre-trained on a large corpus of dialogue data. Additionally, the DialogVED model is based on Bart-large(0.47B), which gives it a significant advantage in terms of the number of parameters compared to our model(0.21B).

The PLATO model uses discrete latent variables, while DialogVED uses a VAE-based continues latent variable, We compared our model with these two models to demonstrate the advantages of handling latent variables using the diffusion model.

To more effectively evaluate the impact of our latent discrete variable, we also conducted a comparison with the version that does not include a latent variable (referred to as 'Our w/o Latent'). It accepts the same context embedding input as our model, and also using Bart-base as it's backbone, sharing the same training settings as our method with latent variables.

The last line represents the upper bound of our model, we generate $10$ different latent variable for the same context and use them to generate corresponding responses as candidates. We select the candidate with the highest overlap with the reference, i.e., the highest Bleu-1 score, as our final result.

DiffusionDialog represents the result of our model with one candidate, and all models use Beam Search for decoding, with a beam size of 5. Our diffusion model utilizes the DDIM\citep{song2020denoising} acceleration technique during inference, with a sampling time step of 50 for the purpose of performance and time efficiency, the results under different inference time steps will be discussed in detail in the later section.


As shown in Table \ref{tab:dailidialog}, our model achieves very high results on the Dist metric. However, compared to the version without latent variable, there is a certain decrease in the Bleu metric, but our model still achieves competitive results in models that have not been pre-trained on dialogue data. The performance of our method without latent variable on the Bleu metric is similar to that of PLATO, which we attribute to the performance of the Bart pre-training model, benefiting from the encoder-decoder architecture and generative pre-training. Compared to DialogVED, which has the same architecture as ours but has more parameters and is pre-trained on dialogue data, our model's Bleu score is much lower.

We notice that the drop in Bleu score due to the introduction of latent variables is smaller on PersonaChat than on DailyDialog. Combining with the statistics in Table \ref{tab:dataset}, we can infer that the data leakage in the test set and development set of DailyDialog penalizes the diversity of generated results.

The improvement in our model's Dist value compared to PLATO and DialogVED indicates that introducing latent variables based on the diffusion model can more effectively improve the diversity of generated responses compared to discrete latent variables and continuous latent variables based on a fixed Gaussian prior. Meanwhile, our experiments on the upper bound of our model's performance also demonstrate the potential of our model.

\subsection{Discussions}

\begin{table}[t]
\centering
\scalebox{1.1}{
\begin{tabular}{c|c}
\hline
Model              & speed           \\ \hline
Our w/o Latent           & 0.068 s/sample  \\ \hline
PLATO              & 25.813 s/sample \\ \hline
DialogVED          & 0.076 s/sample  \\ \hline
Our Method-10$^{\diamond}$    & 0.072 s/sample  \\ \hline
Our Method-100$^{\diamond}$  & 0.189 s/sample  \\ \hline
Our Method-1000$^{\diamond}$ & 1.500 s/sample  \\ \hline
DiffuSeq-10$^{\diamond}$   & 0.384 s/sample  \\ \hline
DiffuSeq-100$^{\diamond}$  & 3.810 s/sample  \\ \hline
DiffuSeq-1000$^{\diamond}$ & 38.246 s/sample  \\ \hline
\end{tabular}}
\caption{Comparison of inference speed among models, models with symbol$^{\diamond}$ utilize diffusion model.}
\label{tab:inference speed}
\end{table}

\subsubsection{Case Analysis}

\begin{table*}[htbp]
\centering
\scalebox{1.0}{
\begin{tabular}{c|cccc|cccc}
\hline
\multirow{2}{*}{Steps} & \multicolumn{4}{c|}{DailyDialog}  & \multicolumn{4}{c}{PersonaChat}   \\
                       & Bleu-1 & Bleu-2 & Dist-1 & Dist-2 & Bleu-1 & Bleu-2 & Dist-1 & Dist-2 \\ \hline
10                     & 0.350  & 0.318  & 0.071  & 0.369  & 0.385  & \textbf{0.331}  & 0.031  & 0.172  \\
100                    & 0.348  & 0.319  & 0.073  & 0.372  & 0.380  & 0.328  & 0.031  & 0.169  \\
1000                   & \textbf{0.352}  & \textbf{0.327}  & \textbf{0.074}  & \textbf{0.373}  & \textbf{0.389}  & \textbf{0.331}  & \textbf{0.032}  & \textbf{0.181}  \\ \hline
\end{tabular}}
\caption{Impact of number of sampling steps on performance.}
\label{tab:diffusion_step}
\end{table*}

In order to further demonstrate the generative capabilities of our model, we provide some generated responses in Table \ref{tab:case}. The table illustrates five of these responses, which showcase the model's ability to generate diverse, relevant, and fluent response.

\subsubsection{Inference Speed}

We compare the inference speed of our model with DialogVED, PLATO, DiffuSeq, and the results are shown in Table \ref{tab:inference speed}.

              


Note that the framwork used for inference among the models are different. PLATO was run on paddlepaddle, DialogVED was run on fairseq, DiffuSeq and our model was just run on pytorch. The number following DiffusionDialog and DiffuSeq represents the number of time steps used for inference.


As the table shows, due to the absence of inference on latent variables, the inference time for our method without latent is very short, and DiffusionDialog is comparable When the number of inference time steps is 10, which demonstrates the high efficiency of our model's inference.

DiffuSeq, like DiffusionDialog, utilizes a diffusion model for text generation. We compare DiffusionDialog with the DiffuSeq model to demonstrate the advantage in inference speed of our diffusion model.

To ensure fairness in comparison, we set the maximum input length of these models to 256. At inference time steps of 10, 100, and 1000, our model required less time for inference than the DiffuSeq model. Moreover, as the number of inference steps increased, our model's speed advantage grew. 

PLATO introduces discrete latent variables, which require generating all candidate responses based on these latent variables, thereby requiring a considerable amount of time. In this comparative experiment, we used 20 discrete latent variables ($K=20$), the same as the official version provided. For DialogVED, we used their large version with $P=64$.

\subsubsection{Sampling Steps}
During inference, the diffusion model requires a large number of sampling steps, which is a significant bottleneck for the inference speed. And prior work, e.g., DiffuSeq\citep{gong2022diffuseq} suffers from a significant drop in generation quality when reducing the sampling steps. In order to investigate the performance of our model on the test dataset under different numbers of sampling steps, we present the results in Table \ref{tab:diffusion_step}.

As shown in the table, our method achieves competitive results with as few
as 10 on both dataset. It should be noted that as the number of sampling steps increases, the performance of our model on PersonaChat, as measured by the BLEU metric, first decreases and then improves. At 1000 time steps, all metrics reach their peak, but the difference between 1000 and 10 steps is not significant.

\section{Related Work}

\subsection{One-to-many Modeling}
The existence of multiple suitable responses for a given context is referred to as the one-to-many problem. Some works introduce latent variable to model the relationship, CVAE\citep{zhao2017learning} utilizes Gaussian distribution to capture variations in responses at the discourse level, since a simple distribution over the latent variables has a lack of granularity in modeling the semantic information of the responses, DialogWAE\citep{gu2018dialogwae} develop a Gaussian mixture prior network to enrich the latent space, instead of the single Gaussian prior of VAE. iVAE$_{\text {MI}}$\citep{fang2019implicit} address the challenge with implicit learning. DialogVED\citep{chen2022dialogved} incorporates continuous latent variables into an enhanced encoder-decoder pre-training framework to increase the relevance and diversity of responses. PLATO\citep{bao2019plato} introduces discrete latent variables to tackle the inherent one-to-many mapping problem in response generation. Both of PLATO and DialogVED are pretrained with large dialog corpus, providing a strong baseline for one-to-many modeling.

\subsection{Diffusion Models for Sequence Learning}
Since Diffusion model\citep{dhariwal2021diffusion, song2020score} has achieved breakthroughs in the field of image processing. There have been many works attempting to apply diffusion models to the field of natural language processing. Considering the discrete nature of texts, D3PM\citep{austin2021structured} introduce Markov transition matrices to diffuse the source data instead of Gaussian noise, Analog Bits\citep{chen2022analog} represents discrete data as binary bits, and then training a continuous diffusion model to model these bits as real numbers. Diffusion-LM\citep{li2022diffusion} develop a non-autoregressive language model based on continuous diffusions with an embedding function and rounding process, iteratively denoises a sequence of Gaussian vectors into words. DiffuSeq\citep{gong2022diffuseq} propose a diffusion model designed for sequence-to-sequence text generation tasks utilizing encoder-only Transformers. And SeqDiffuSeq\citep{yuan2022seqdiffuseq} approach sequence-to-sequence text generation with Encoder-Decoder Transformers. LD4LG\citep{lovelace2022latent} learn the continuous diffusion models in the latent space of a pre-trained encoder-decoder model.
\section{Conclusion}
This paper presents DiffusionDialog, which combines an encoder-decoder structured pre-trained language model with diffusion model. By utilizing the diffusion model to learn the latent space and infer the latent by denoising step by step, we greatly enhance the diversity of dialog response while keeping the coherence and achieving high inference efficiency. As experimental results shows, our model has achieved a over 50\% increase in the dist metric and accelerate inference speed over 50 times compared to the DiffuSeq model. Overall, this work provides a novel idea for applying diffusion model into natural language processing.

\section{Limitations}
As shown in the experiments, the accuracy of our model is not yet high enough. We identified two main reasons for this: 1) we have not conducted extensive pre-training, and 2) the structure and training methods of the model are not yet optimal. We will attempt to address these issues in future work.

\section{Acknowledgments}

This work is supported by the National Natural Science Foundation of China (Grant No. 62261160648 and 62376177). This work is also supported by Collaborative Innovation Center of Novel Software Technology and Industrialization, the Priority Academic Program Development of Jiangsu Higher Education Institutions, and the joint research project of Meituan and Soochow University. We would also like to thank the anonymous reviewers for their insightful and valuable comments.

\nocite{*}
\section{References}\label{sec:reference}

\bibliographystyle{lrec-coling2024-natbib}
\bibliography{custom}


\end{document}